\pdfoutput=1

\documentclass[11pt]{article}

\usepackage[preprint]{acl}

\usepackage{times}
\usepackage{latexsym}

\usepackage[T1]{fontenc}

\usepackage[utf8]{inputenc}

\usepackage{microtype}

\usepackage{inconsolata}

\usepackage{graphicx}
\usepackage{multirow}
\usepackage{amssymb}
\usepackage{amsmath}  
\usepackage{booktabs}

\usepackage{wrapfig}
\usepackage{graphicx}
\usepackage{booktabs}
\usepackage{subfigure}
\usepackage{placeins} 
\usepackage{verbatim}
\usepackage{fancyvrb}
\usepackage{fvextra} 
\usepackage[most]{tcolorbox}
\usepackage{float} 
\lstdefinestyle{mystyle}{
    columns=fullflexible,
    backgroundcolor=\color{backcolour},   
    commentstyle=\color{codegreen},
    numberstyle=\tiny\color{codegray},
    stringstyle=\color{codepurple},
    basicstyle=\footnotesize,
    breakatwhitespace=false,         
    breaklines=true,  
    breakindent=0pt,
    captionpos=b,                    
    keepspaces=true,  
    numbersep=0pt,                  
    showspaces=false,                
    showstringspaces=false,
    showtabs=true,                  
    tabsize=10,
    escapeinside={@}{@}, 
    moredelim=**[is][\highlight]{`}{`}, 
}
\definecolor{ngreen}{HTML}{76B900}

\usepackage{xspace} 
\newcommand{\ouralgo}{Socratic-MCTS\xspace}

\newif\ifshowcomments

\showcommentsfalse
\ifshowcomments
  \newcommand\hyunwoo[1]{{\color{red}[#1]$_{HW}$}}
  \newcommand\ximing[1]{{\color{purple}[#1]$_{Ximing}$}}
  \newcommand\jaehun[1]{{\color{olive}[#1]$_{Jaehun}$}}
  \newcommand\amlan[1]{{\color{orange}[#1]$_{Amlan}$}}
  \newcommand{\DA}[1]{{\color{red}{[David: #1]}}}
  \newcommand\yejin[1]{{\color{cyan}[#1]$_{yejin}$}}
\else
  \newcommand\hyunwoo[1]{}
  \newcommand\ximing[1]{}
  \newcommand\jaehun[1]{}
  \newcommand\amlan[1]{}
  \newcommand{\DA}[1]{}
  \newcommand\yejin[1]{}
\fi

\newcommand{\model}{\mathcal{M}}           
\newcommand{\question}{q}                  
\newcommand{\Image}{I}  
\newcommand{\questionspace}{\mathcal{Q}}   
\newcommand{\answer}{a}                    
\newcommand{\answerspace}{\mathcal{A}}     

\newcommand{\reasoning}{T}                 
\newcommand{\reasoningspace}{\mathcal{T}}  

\newcommand{\thought}[1]{u^{#1}}           
\newcommand{\subquestion}[1] {s_{#1}}
\newcommand{\answersubq}[1] {{a}_{s#1}}
\newcommand{\transitionwrapup}{\textnormal{wrap}}

\newcommand{\prompt}[1]{\textnormal{p}_{\textnormal{#1}}}
 
\newcommand{\node}[1]{\mathbf{n}_{#1}}
\DeclareMathOperator*{\argmax}{arg\,max}

\title{Socratic-MCTS:\\ Test-Time Visual Reasoning by Asking the Right Questions}

\author{
David Acuna$^\dagger$\textsuperscript{\textcolor{ngreen}{$\clubsuit$}} \quad
Ximing Lu$^\dagger$$^\ddagger$\textsuperscript{\textcolor{ngreen}{$\clubsuit$}} \quad
Jaehun Jung$^\dagger$$^\ddagger$\textsuperscript{\textcolor{ngreen}{$\clubsuit$}} \quad
Hyunwoo Kim$^\dagger$\textsuperscript{\textcolor{ngreen}{$\clubsuit$}} \\
\textbf{Amlan Kar}$^\dagger$$^\mathsection$ \quad
\textbf{Sanja Fidler}$^\dagger$$^\mathsection$ \quad
\textbf{Yejin Choi}$^\dagger$ \\[0.2cm]
$^\dagger$NVIDIA \quad
$^\ddagger$University of Washington \quad
$^\mathsection$University of Toronto\\[0.2cm]
\texttt{\{dacunamarrer, ximingl, jaehunj, hyunwook, yejinc\}@nvidia.com}
}

\begin{document}
\maketitle

\begingroup
\renewcommand\thefootnote{\textcolor{ngreen}{$\clubsuit$}}
\footnotetext{\hspace{0.1cm} First co-authors.}
\endgroup

\begin{abstract}

Recent research in vision-language models (VLMs) has centered around the possibility of equipping them with implicit long-form chain-of-thought reasoning—akin to the success observed in language models—via distillation and reinforcement learning. But what about the non-reasoning models already trained and deployed across the internet? Should we simply abandon them, or is there hope for a search mechanism that can elicit hidden knowledge and induce long reasoning traces—\textit{without} any additional training or supervision?
In this paper, we explore this  possibility using a Monte Carlo Tree Search (MCTS)-inspired algorithm, which injects subquestion–subanswer pairs into the model’s output stream.  
We show that framing reasoning as a search process—where subquestions act as latent decisions within a broader inference trajectory—helps the model “connect the dots” between fragmented knowledge and produce extended reasoning traces  in non-reasoning models.
We evaluate our method across three benchmarks and observe consistent improvements. Notably, our approach yields a 2\% overall improvement on MMMU-PRO, including a significant 9\% gain in Liberal Arts. 

\end{abstract}

\begin{figure}[t!]
\vspace{-0.5cm}

    \includegraphics[width=0.48\textwidth]{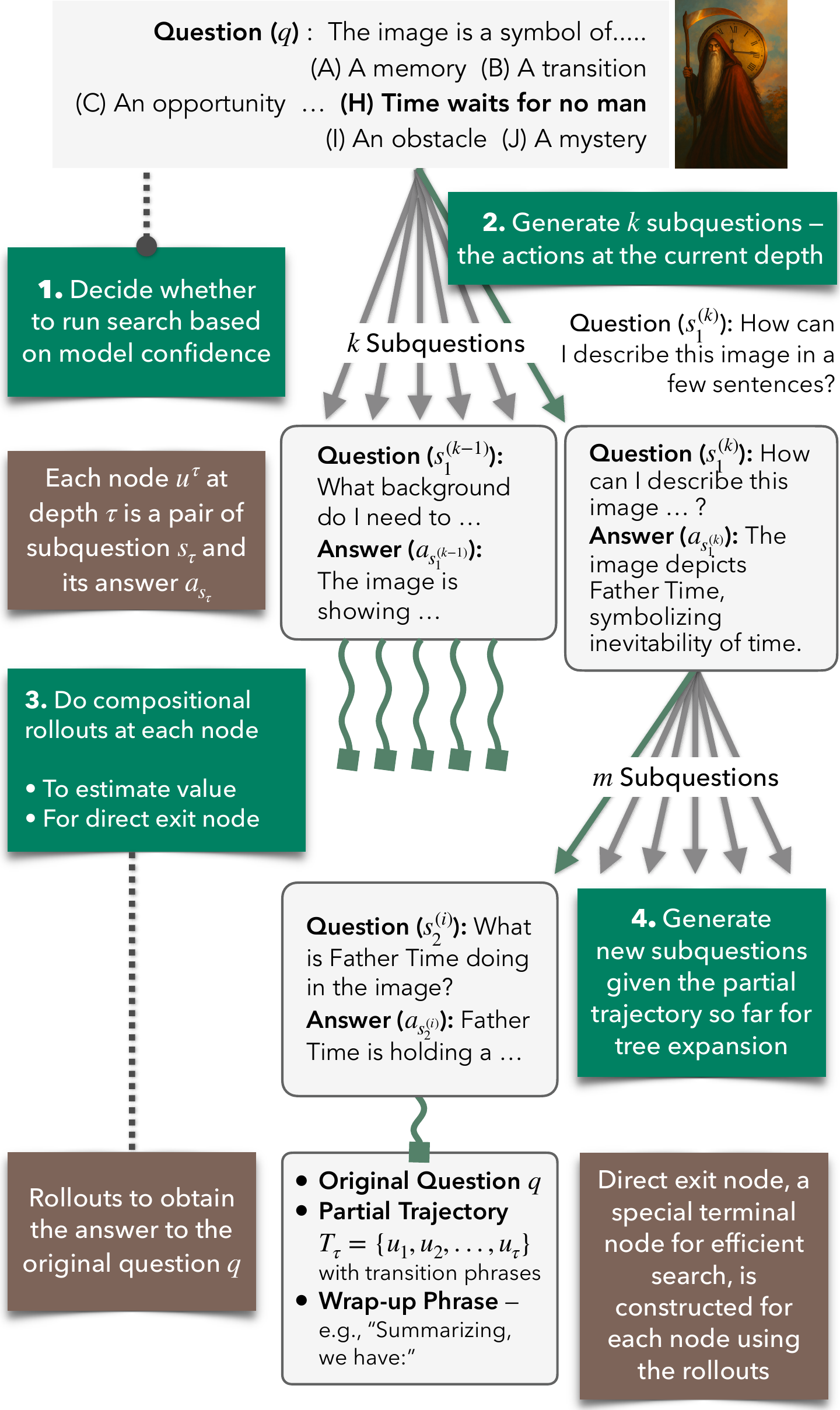}
    \vspace{-7.8mm}
    \caption{
\textbf{Socratic-MCTS Overview.} In Socratic-MCTS, \emph{actions} are defined as \emph{subquestions}, and each \emph{node state} consists of a \emph{subquestion–subanswer} pair. During search, rollouts are performed by preconditioning the model on the accumulated reasoning trajectory in a compositional manner. To structure this trajectory and enable faster rollouts, we use \emph{transition phrases} (e.g., “First, I need to consider...”) and conclude with a \emph{wrap-up phrase} (e.g., “Summarizing, we have:”), which cue the model to complete the reasoning and produce a final answer. We estimate value through internal agreement and incorporate early-exit and selective search mechanisms to adaptively reduce computational overhead—all without external supervision.
}

    \label{fig:your_figure_label}
    \vspace{-7.5mm}
\end{figure}%

\begin{figure*}[t!]
\centering

\subfigure[Socratic-MCTS on a logical reasoning question.]{
    \includegraphics[width=0.47\textwidth]{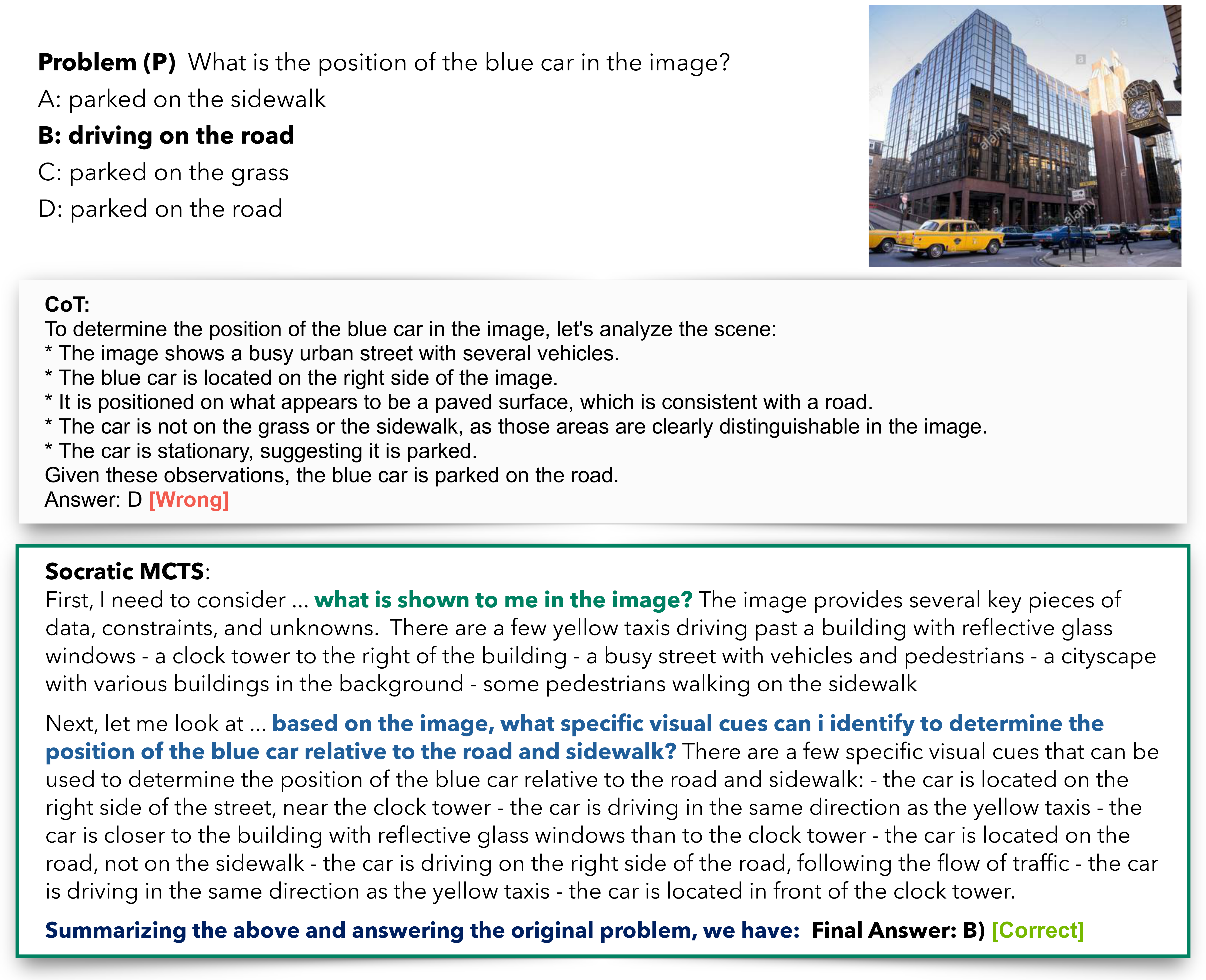}
    \label{fig:mmstar1}
}
\hfill
\subfigure[Socratic-MCTS on a chart  question.]{
    \includegraphics[width=0.47\textwidth]{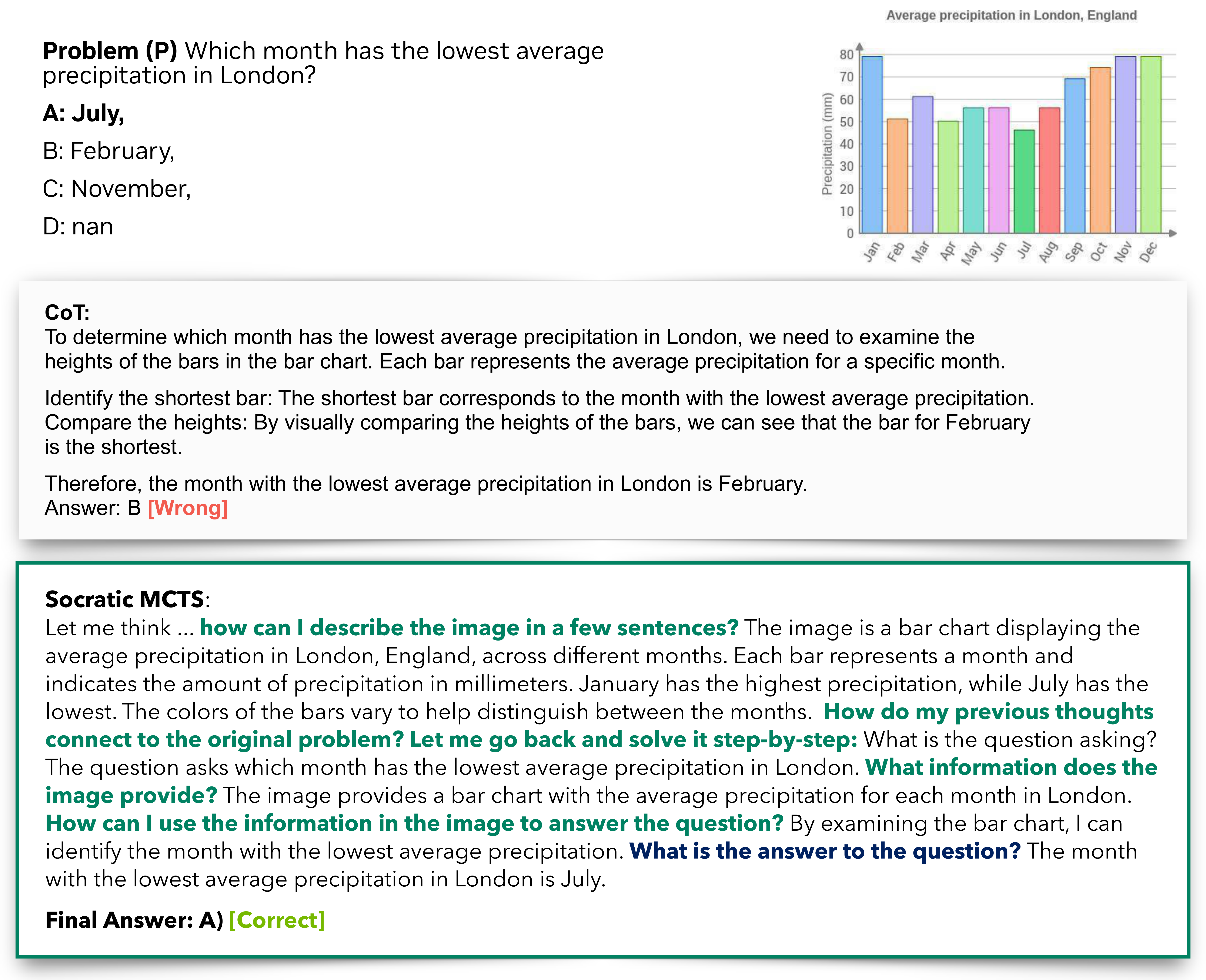}
    \label{fig:mmstar2}
}

\caption{\textbf{Qualitative comparison on MMStar.} We show Socratic-MCTS responses on two multimodal questions from the benchmark, comparing qualitatively against the \textit{CoT prompting} baseline. Socratic-MCTS allows the model to uncover relevant knowledge, verify intermediate steps, and synthesize final answers coherently.}
\label{fig:mmmu-qual-comparison}
\end{figure*}

\section{Introduction}

Recent state-of-the-art large language models (LLMs) have demonstrated remarkable capabilities in complex reasoning tasks~\citep{chen2024sharegpt4v,deepseekai2025deepseekr1incentivizingreasoningcapability,openai2024_o1,liu2025prorl}. A key driver of this progress has been the emergence of long chain-of-thought (CoT) reasoning through post-training with techniques such as reinforcement learning and distillation~\citep{deepseekai2025deepseekr1incentivizingreasoningcapability,lu2025retro,prismaticsynthesis,muennighoff2025s1simpletesttimescaling}. Inspired by this success, much of the recent focus in vision-language modeling has shifted toward mirroring these advances, attempting to induce reasoning capabilities primarily through reinforcement learning and distillation~\citep{du2025virgopreliminaryexplorationreproducing,liao2025longperceptualthoughts}.

But what about the many existing non-reasoning VLMs already trained and publicly available? As of this writing, the vast majority of open-weight VLMs fall into this category~\citep{chen2024allava,liao2025longperceptualthoughts,bai2025qwen25vltechnicalreport}. Should we simply abandon them, or is there hope for mechanisms that can repurpose these models—eliciting structured reasoning without additional training or supervision? While at first glance this may seem unfeasible, these models have been trained on vast internet-scale data and may possess latent knowledge 
and reasoning potential that conventional prompting fails to activate. Can we unlock reasoning in non-reasoning VLMs by eliciting hidden, belated knowledge 
inaccessible through standard "step-by-step" CoT prompting?

In this work,  we
propose \textbf{\ouralgo}, a test-time algorithm that frames reasoning in non-reasoning models as a structured search problem—requiring no fine-tuning, no supervision, and no architectural modifications. At its core, \ouralgo\ introduces a simple but powerful abstraction: the subquestion–subanswer pair.
By framing reasoning as a search process where subquestions represent latent decisions within a broader inference trajectory, we enable models to “connect the dots” between fragmented pieces of knowledge. By searching over semantically meaningful chunks, rather than individual lines, \ouralgo\ strikes a middle ground between free-form generation and conventional tree search—producing coherent long reasoning traces that progressively move toward the final solution. Our key insight is that longer CoT could emerge at test-time from the deliberate exploration of which subquestions to ask, and in what order—injecting structure into the model’s output stream to guide it toward correct answers.
\ouralgo\ formalizes this process through an MCTS-inspired framework 
that 
estimates 
and incorporates early-exit mechanisms to adaptively reduce computational overhead
—all without relying on external supervision.
We evaluate \ouralgo\ across three benchmarks and find that it consistently improves performance, notably achieving a significant 9\% gain in Liberal Arts categories on the  MMMU-PRO benchmark.

\section{Preliminaries}


Consider a model $\model$ that, when 
given a question $\question := \{\Image, \hat{\question}\}$, where $\Image$ is an image and $\hat{\question}$ is a text prompt of a multiple-choice question, generates both an intermediate reasoning \textbf{trajectory} $\reasoning$ and a final solution $\answer$.  
Formally, given an input question $\question \in \questionspace$, the model $\model$ produces a pair $(\reasoning, \answer) := \model(\question)$, where $\reasoning \in \reasoningspace$ denotes the chain of reasoning (or “thoughts”), and $\answer \in \answerspace$ denotes the final answer to $\question$.  
Each reasoning trajectory $\reasoning$ consists of a \textbf{set of core units of reasoning}, denoted $\reasoning := \{\thought 1, \thought 2, \ldots, \thought \tau\}$. Each $\thought \tau$ represents an individual reasoning step.  
In most settings, $\thought \tau$ corresponds to a sentence or logical operation—e.g., a mathematical step—and is often delineated in model outputs via sentence colons.
In our framework, each atomic reasoning unit is represented as $\thought \tau := \{\subquestion \tau, \answersubq \tau\}$, where $\subquestion \tau$ is a sub-question and $\answersubq \tau$ is the corresponding answer. 
To facilitate segmentation between reasoning units, we  append  \textbf{transition phrases}—natural language cues such as “First, let me think …”
at the beginning of each thought $\thought i$. These  
structure 
the trajectory allowing coherent composition. 

\noindent\textbf{Monte Carlo Tree Search}
Monte Carlo Tree Search (MCTS) is a sample‑based tree-search algorithm 
that 
has proved effective on complex reasoning tasks.
In our case, MCTS operates on a rooted, directed tree whose nodes correspond to partial reasoning trajectories.  
We represent a node at depth~$\tau$ as $
  \node \tau := \bigl\langle \question,\; \reasoning_\tau,\; N,\; W,\; Q \bigr\rangle $
where  
(i)~$\question$ is the original query,  
(ii)~$\reasoning_\tau=\{\thought 1,\dots,\thought \tau\}$ is the sequence of core reasoning units generated so far (with $\thought 0:=\varnothing$ for the root),  
(iii)~$N\in\mathbb{N}$ is the visit count,  
(iv)~$W\in\mathbb{R}$ is the cumulative reward returned by completed roll‑outs that passed through the node, and  
(v)~$Q:=W/N$ is its empirical value estimate.  
An edge from $\node \tau$ to $\node {\tau+1}$ is defined by an action.
MCTS operates in four iterative phases: selection, expansion, simulation, and backpropagation.  
These  are repeated until a predefined computational budget is reached.  
We refer the reader to \citet{browne2012survey} for details. 
\section{Socratic MCTS}

%
\label{sec:socratic_mcts}

We adapt MCTS to a \emph{Socratic} setting in which
\emph{actions} are \textbf{subquestions} 
and each \textbf{node state} is the pair
$\thought \tau := \{\subquestion \tau,\answersubq \tau\}$ comprising the current
subquestion and its answer.
This is inspired by the Socratic method of reasoning, as the model explicitly and progressively decomposes the problem by asking and answering intermediate questions. Thus, unlike previous MCTS approaches that implicitly sample tokens or steps as actions, we explicitly define subquestions as structured dynamic actions, aligning MCTS with the Socratic method~\cite{farnsworth2021socratic}.
We emphasize that our algorithm operates  at test time and does not rely on supervision. Instead, it leverages compositional rollouts and internal agreement to guide the search. 
\subsection{Explicit Subquestions as Actions}
Previous work on applying MCTS to LLMs typically leaves the action space
implicit—the next token or thought sampled from the model
\citep{guan2025rstar,omegaprm}.  
Instead, we define an action at depth~$\tau$ as a
self‑contained subquestion $\subquestion {\tau+1}$ that (i) relates to the
original query $\question$ and (ii) decomposes the task into a tractable
sub‑problem.  
This explicit representation allows searching
over semantically meaningful  pieces of knowledge
while
preserving goal‑directedness. 
Specifically,  we obtain a finite number of subquestions $k_q$:
$
  \bigl\{\subquestion {\tau+1},\dots
  ,\subquestion {\tau+k_q}\bigr\}
  \;\sim\;
  \model_s\!\bigl(\cdot \mid \question,\reasoning_\tau,\prompt {sub}\bigr),
$
where $\model_s$ is the \emph{subquestion policy} and the prompt $\prompt {{sub}}$ instructs the model to \emph{ask} rather than \emph{answer}.
%
Each sampled subquestion defines an edge from $\node \tau$ to a new child $\node {\tau+1}$.  The corresponding answer is obtained from an
\emph{answer policy}:
$
  \answersubq {\tau+1} 
  \;\sim\;
  \model_a\!\bigl(\cdot \mid \Image, \subquestion {\tau+1}\bigr),
$
which answers the subquestion
\emph{in isolation}.  Empirically, we found this decoupling crucial in multimodal models, preventing propagating errors or contaminating the answer.
The new node’s state is then
$\thought {\tau+1}:=\{\subquestion {\tau+1},\answersubq {\tau+1}\}$.
%
We make the transition from $\thought \tau$ to $\thought {\tau +1}$ natural by appending a transition phrase such as "Next, let me ..." drawn from a list.
For simplicity, we set $\model_s = \model_a = \model$ in our experiments, and set $temp=0.6$, deferring the exploration of heterogeneous or multi-agent policies within the Socratic MCTS framework to future work. Notably, any agent—including a different model or a human—could be placed in either role.

\subsection{Navigating the Reasoning Tree}

\noindent\textbf{Guided Selection via UCT.}
Starting at the root $\node 0$, we repeatedly choose the child that maximises
$
  \textnormal{UCT}(\node {\tau+1})
  \;=\;
  Q \;+\; c\,\sqrt{\frac{\ln N_{\mathrm{parent}}}{N_{\tau+1}}}\!,
$
until a leaf is reached.  The constant $c>0$ controls the
exploration–exploitation balance. Following common practice~\citep{kocsis2006bandit}, we set the exploration constant $c=1.4$.

\noindent\textbf{Expanding with Socratic Questions.}
If the current leaf node is non-terminal and not fully expanded, we generate up to $k_q$ new sub-questions and their answers as described above, initializing each child with $N=W=0$. A node is considered \emph{fully expanded} when all of its sampled subquestions have been explored (i.e., all available actions have been taken). A node is considered \emph{terminal} if one of two conditions holds: (1) the model fails to generate an answer for a proposed action (e.g., empty answer), or (2) the generation degenerates (e.g., repetitive outputs). In all  experiments, $k_q=6$ for the first tree level and $k_q=3$ for the rest.

\noindent\textbf{Compositional Rollouts for Self-consistency.}
We perform rollouts to estimate a node’s value by composing transition phrases, subquestions, and their answers along the reasoning path.
Concretely, 
we precondition the model on the current partial trajectory $\reasoning_{\tau+1}$.
We additionally concatenate a \textit{wrap-up transition phrase} (e.g., “Summarizing, we have:”)
that cues the model to complete the reasoning and produce a final answer.
Formally, we generate $K$ completions by preconditioning the model on $\reasoning_{\tau+1}$ followed by $K$ distinct wrap-up transition phrases:
$
  \answer^{(1)}, \dots, \answer^{(K)}
  \;\sim\;
  \model\bigl(\cdot \mid \question,\reasoning_{\tau+1}, \transitionwrapup^{(k)}\bigr),
$
where $\transitionwrapup^{(k)}$ denotes the $k$-th natural language wrap-up phrase. 
Notably, this procedure enables efficient rollouts—requiring only a few newly generated tokens—as the model is preconditioned to produce a final response.
We also observe that diverse wrap-up phrases yield variability in the response, which is necessary for computing internal agreement.
We set $K=8$  in all our experiments.  

\noindent\textbf{ Internal Agreement for Value Estimation.}
We rely on the model's internal consistency as a proxy reward signal.  
Specifically, each of the sampled answer $\answer^{(k)}$ is parsed using a lightweight heuristic that extracts a canonical choice (e.g., a multiple-choice label), and penalizes overly verbose generations.
Formally, let the set of unique extracted answers from all rollouts be $\mathcal{A} := \{\hat{a}^{(1)}, \dots, \hat{a}^{(K)}\}$, and let $w^{(k)} := \mathrm{score}(\answer^{(k)}) \in [0,1]$ be a normalized confidence weight assigned by the heuristic.
For example, if $\answer^{(k)}$ is degenerate—i.e., it reaches the maximum generation length without producing a valid answer—then its weight $w^{(k)}=0$.
We compute the value estimate via weighted majority voting:
$
  V := \argmax_{a \in \mathcal{A}} \sum_{k=1}^K \mathbf{1}\big[\hat{a}^{(k)} = a\big] \cdot w^{(k)}.
$
That is, we select the answer that accumulates the highest total confidence across the $K$ rollouts.  

\paragraph{Special Direct-Exit Nodes.}
While we encourage the algorithm to explore deeper chains of reasoning when beneficial, we also want to allow it to finalize an answer early if further composition fails to improve the value estimate.
To support this, we allow each node to optionally include a \emph{direct exit node}. 
This is a special terminal node that includes the terminal answer $\answer$ in the reasoning trajectory—obtained from compositional rollouts at that node as explained above.   
To
mitigate an unintended bias introduced by this special node, we omit the exploration term when computing the UCT score on them.
%

\noindent\textbf{Selective Search Based on Model Confidence.} 
We observe that for certain problems, the model exhibits high confidence in its initial direct answer obtained through vanilla sampling.
To handle these cases efficiently, we first estimate the model's confidence in its initial answer and introduce a hyperparameter-controlled early-exit threshold that skips the tree-search algorithm altogether when exceeded (we used 0.9).
We explore two approaches for confidence estimation: taking the maximum answer token probability and performing majority voting through sampling. Empirically, we did not observe a major difference between them.  Future work might explore other adaptive schemes.

\begin{table}[t]
    \centering
    \resizebox{0.5\textwidth}{!}{%
    \begin{tabular}{l|ccc|c}
    \toprule
    \textbf{Method} & 
    \multicolumn{3}{c|}{\textbf{MMMU-Pro}} & 
    \textbf{MMStar} \\
    & Liberal Arts & STEM+B & Overall & Overall \\
    \midrule
    GPT-4o (051324) & - & - & 0.540 & 0.647 \\
    Claude 3.5-Sonnet & - & - & 0.550 & 0.651 \\
    Gemini-1.5-Pro & - & - & 0.494 & 0.591 \\
    LLaVA-OneVision-72B & - & - & 0.380 & 0.658 \\
    Qwen2-VL-72B & - & - & 0.492 & 0.683 \\
    InternVL2-Llama3-76B & - & - & 0.419 & 0.674 \\
    \midrule
    \textbf{InternVL2.5-78B} & & & & \\
    + Direct & 0.538 & \textbf{0.507}  & 0.517 & 0.692 \\
    + CoT & 0.544 & 0.479 & 0.506 & 0.689 \\
    + Least-to-Most & 0.296 & 0.276 & 0.280 & 0.486 \\
    + \textbf{Socratic MCTS (Ours)} & \textbf{0.628} & {0.492} & \textbf{0.537} & \textbf{0.711} \\
    \bottomrule
    \end{tabular}
    }
    \caption{Performance across different reasoning benchmarks. Socratic MCTS consistently outperforms direct, CoT and LtM baselines across all benchmarks. We evaluate  InternVL-78B Direct, CoT, LtM. Others results are for reference, unknown entries represented as -.  }
    \label{tab:main-results}
\end{table}

\begin{figure*}[t!]
\centering

\subfigure[Socratic-MCTS on a multimodal color theory question.]{
    \includegraphics[width=0.47\textwidth]{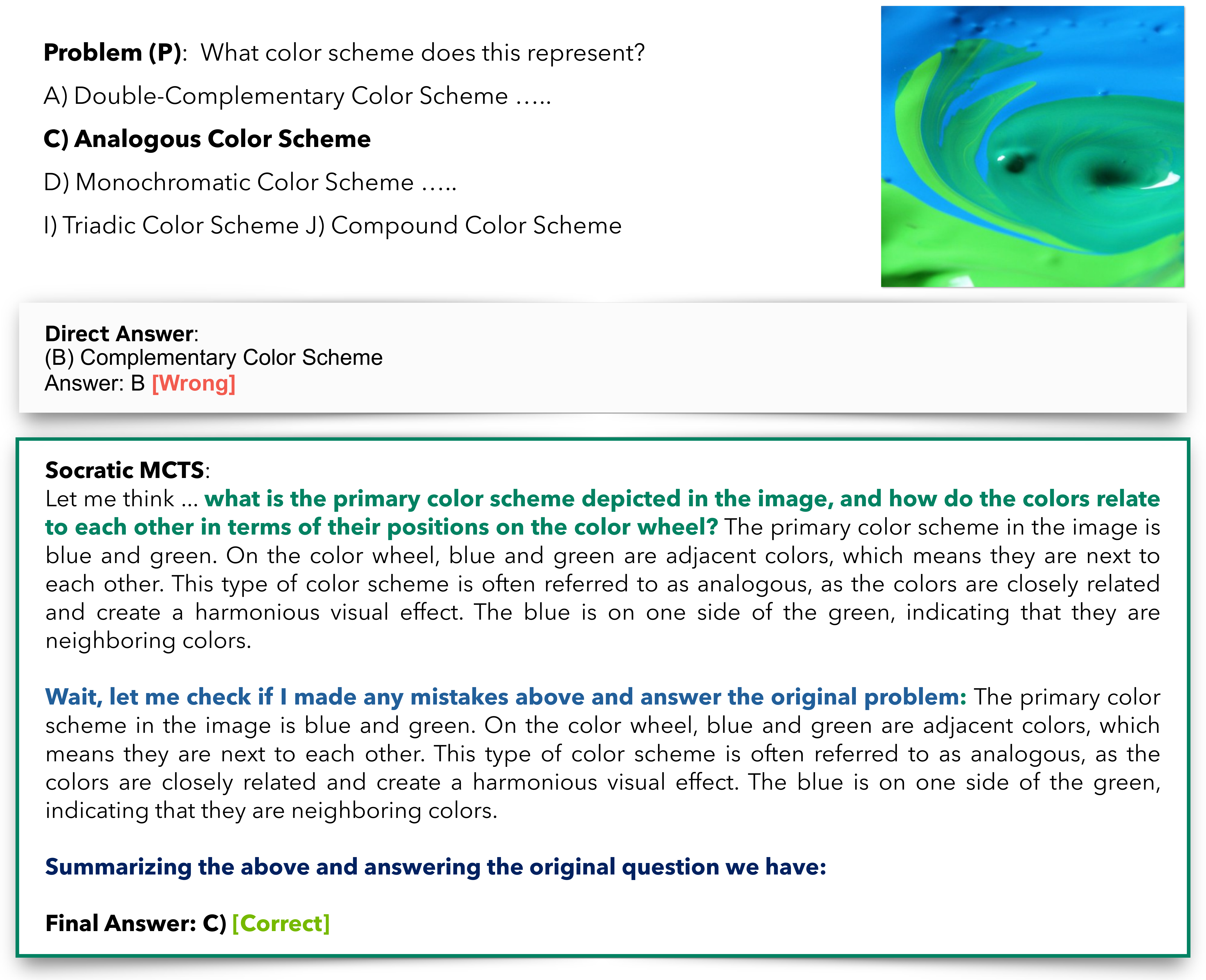}
    \label{fig:mmmu-color}
}
\hfill
\subfigure[Socratic-MCTS on a music question.]{
    \includegraphics[width=0.47\textwidth]{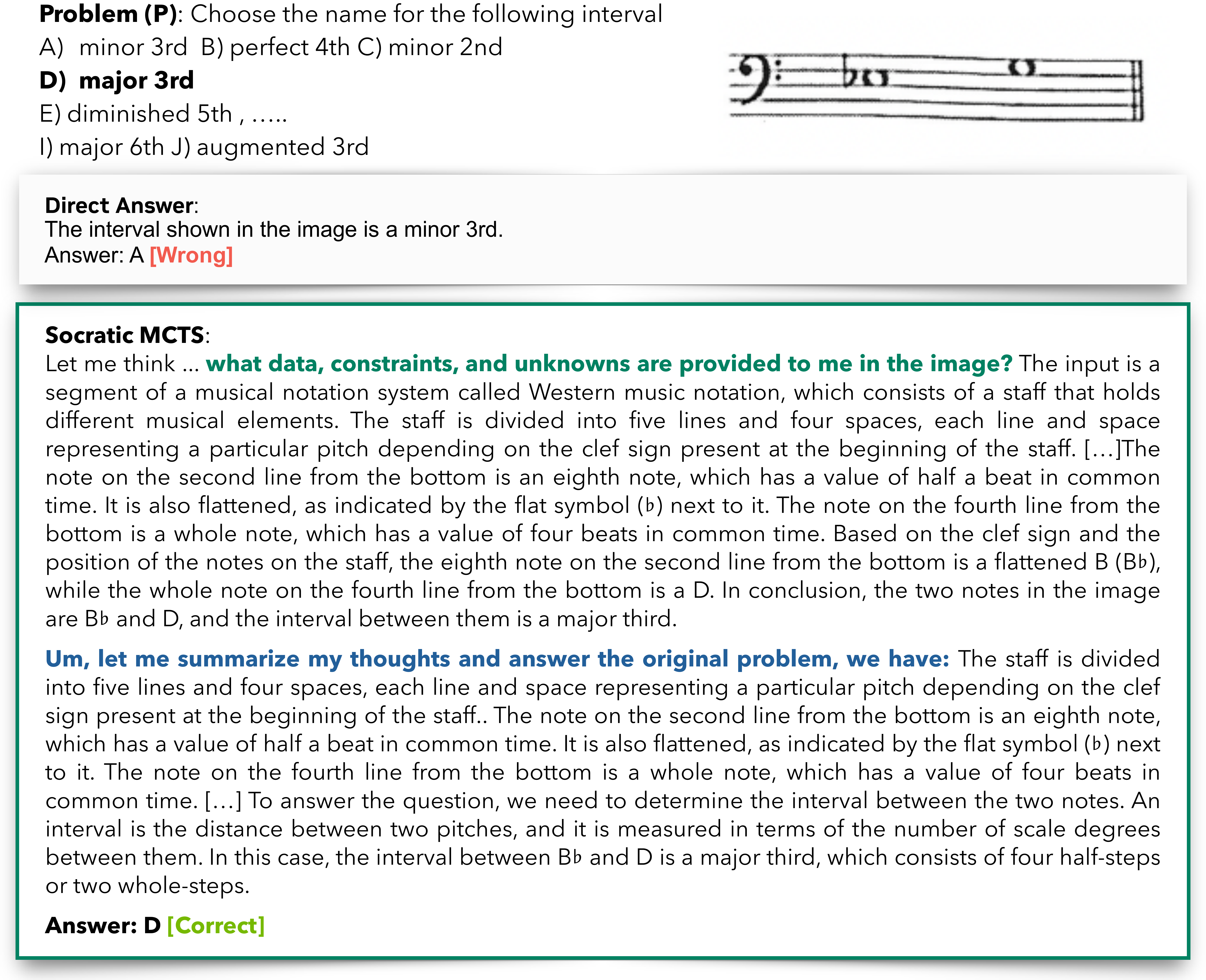}
    \label{fig:mmmu-music}
}

\caption{\textbf{Qualitative comparison on MMMU-PRO.} We show Socratic-MCTS responses on non-symbolic tasks comparing qualitatively against the best-performing baseline in this benchmark (\textit{direct prompting}). }
\label{fig:mmmu-qual-comparison}
\end{figure*}

\section{Experiments}

\textbf{Setup}. We evaluate InternVL-78B~\cite{chen2024expanding} using default hyperparams, and compare it against non-reasoning SoTA models GPT-4o (240513) and Claude 3.5-Sonnet, with results from~\cite{chen2024expanding}. We benchmark three baselines: (1) Direct Answer, (2) CoT prompting, using specific prompts from~\cite{mmmupro} and greedy decoding. We further implemented a multimodal version of Least-To-Most (LtM). For LtM, we recursively decompose the original question up to a depth of 3 and solve the resulting sub-questions in a bottom-up fashion. We evaluate on MMMU-PRO~\cite{mmmupro}, a more challenging variant of MMMU~\cite{mmmu}, under a 10-option multiple-choice format. We also assess performance on MMStar~\cite{chen2024we} (1,500 visually grounded, leakage-controlled samples) and MathVista~\cite{lu2023mathvista} (English-only, multiple-choice subset of \texttt{test-mini}) for visual math  reasoning. 
We report single runs with 40 iterations for MMMU-PRO and 20 for others.

\noindent\textbf{Main Results}. 
\ouralgo\ consistently outperforms all baselines~(Tab~\ref{tab:main-results} and \ref{tab:mathvista-results}) 
, with  strong gains in tasks requiring less symbolic reasoning. In MMMU-Pro, it notably improves accuracy by ~9\% in Liberal Arts.
We define \textit{Liberal Arts} as Art, Art Theory, Design, Economics, Geography, History, Literature, Music, Psychology, and Sociology, and  all remaining subjects as  \textit{STEM+B}.
Surprisingly, across all benchmarks, we find that in VLMs, decomposition via prompting —as in LtM— underperforms both direct and CoT, underscoring the limitations of prompting  and the fundamental differences between non-reasoning VLMs and LLMs.

\begin{table}
\centering
\small
\resizebox{0.4\textwidth}{!}{%
\begin{tabular}{l|c}
\toprule
\textbf{Method} & \textbf{MathVista (mini-eng)} \\
\midrule
+ Direct & 0.740 \\
+ CoT & 0.763 \\
+ Least-to-Most & 0.471 \\
+ \textbf{Socratic MCTS (Ours)} & \textbf{0.782} \\
\bottomrule
\end{tabular}
}
\caption{Performance on MathVista mini further filtered to include only multiple-choice
questions in English.}\label{tab:mathvista-results}
%
\end{table}

%

%

\section{Related Work}

%
Prior work has explored question decomposition for LLMs as a direct prompting method~\citep{radhakrishnan2023question,zhou2022least,khot2022decomposed,jung2022maieuticpromptinglogicallyconsistent,liao2024feedbackenhancesemanticgrounding,liao-etal-2024-reasoning}, mainly in text-only domains. In contrast, we integrate decomposition into a tree search process, allowing models to explore and compose subquestions dynamically, with a focus on vision. 
Recent adaptations of MCTS to VLMs~\citep{yao2024mulberry,wu2025boosting} 
either use MCTS as part of a training loop
or in a small dataset to generate high-level reasoning templates.
Closer to our setting, \citet{hao2023reasoning} use MCTS and frame subquestions as actions in math tasks. Our work differs in both domain and methodology: we target non-reasoning VLMs, introduce preconditioning in rollouts and early-exit mechanisms, and operate in the multimodal domain—producing long  reasoning traces that require grounded visual understanding.

%

\section{Conclusions}

We introduce \ouralgo, a test-time algorithm that frames reasoning as structured search over subquestion–subanswer pairs.
Evaluations across multiple benchmarks show consistent performance gains, particularly on non-symbolic tasks—validating that fragmented knowledge can be elicited to produce long reasoning traces in “non-reasoning” VLMs, without additional training.


\begin{figure*}[h]
\centering

\FloatBarrier
\subfigure[Socratic-MCTS failure case on MMStar.]{
    \includegraphics[width=0.47\textwidth]{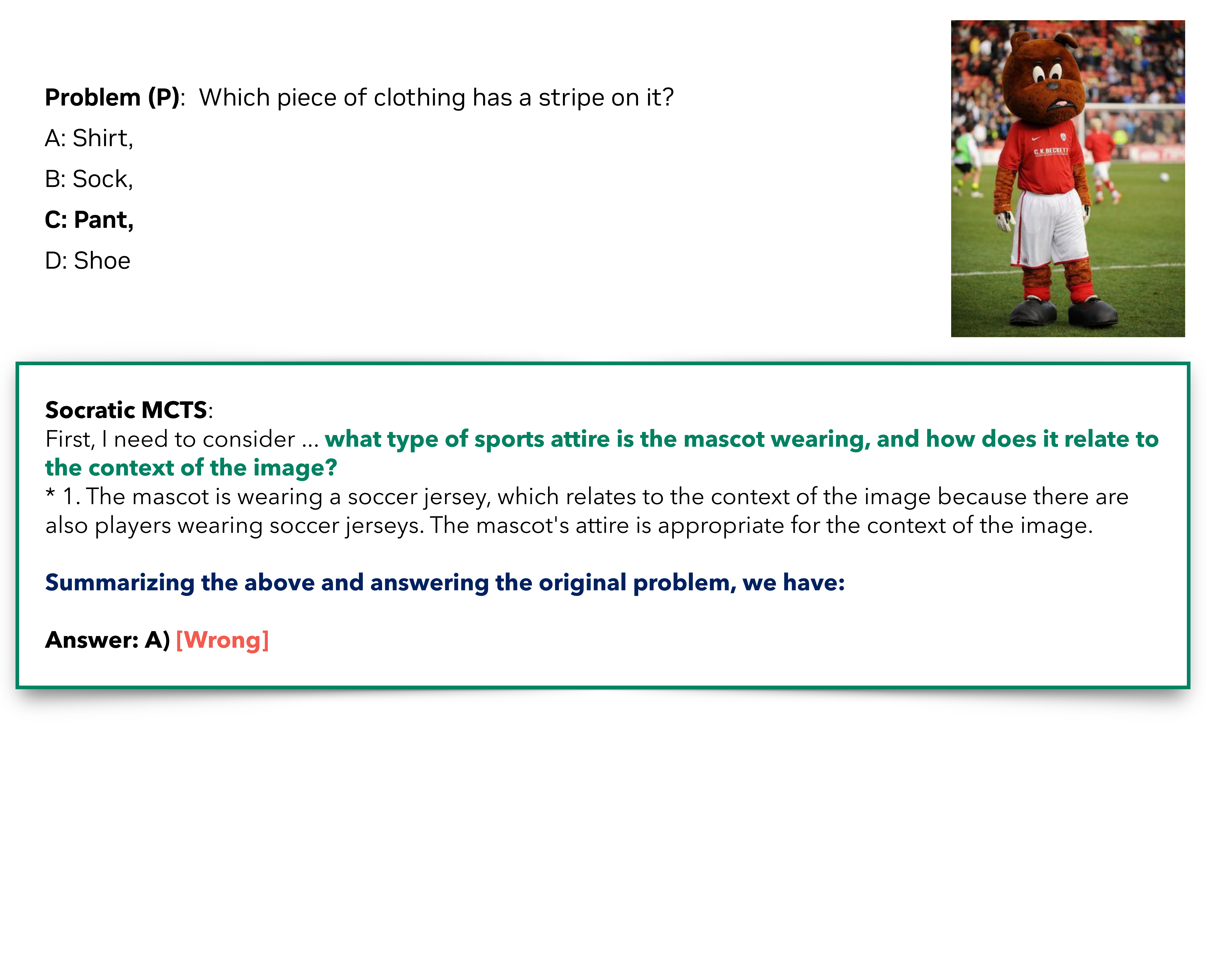}
    \label{fig:mmstar_fail}
}
\hfill
\subfigure[Socratic-MCTS failure case on MMMU-Pro]{
    \includegraphics[width=0.47\textwidth]{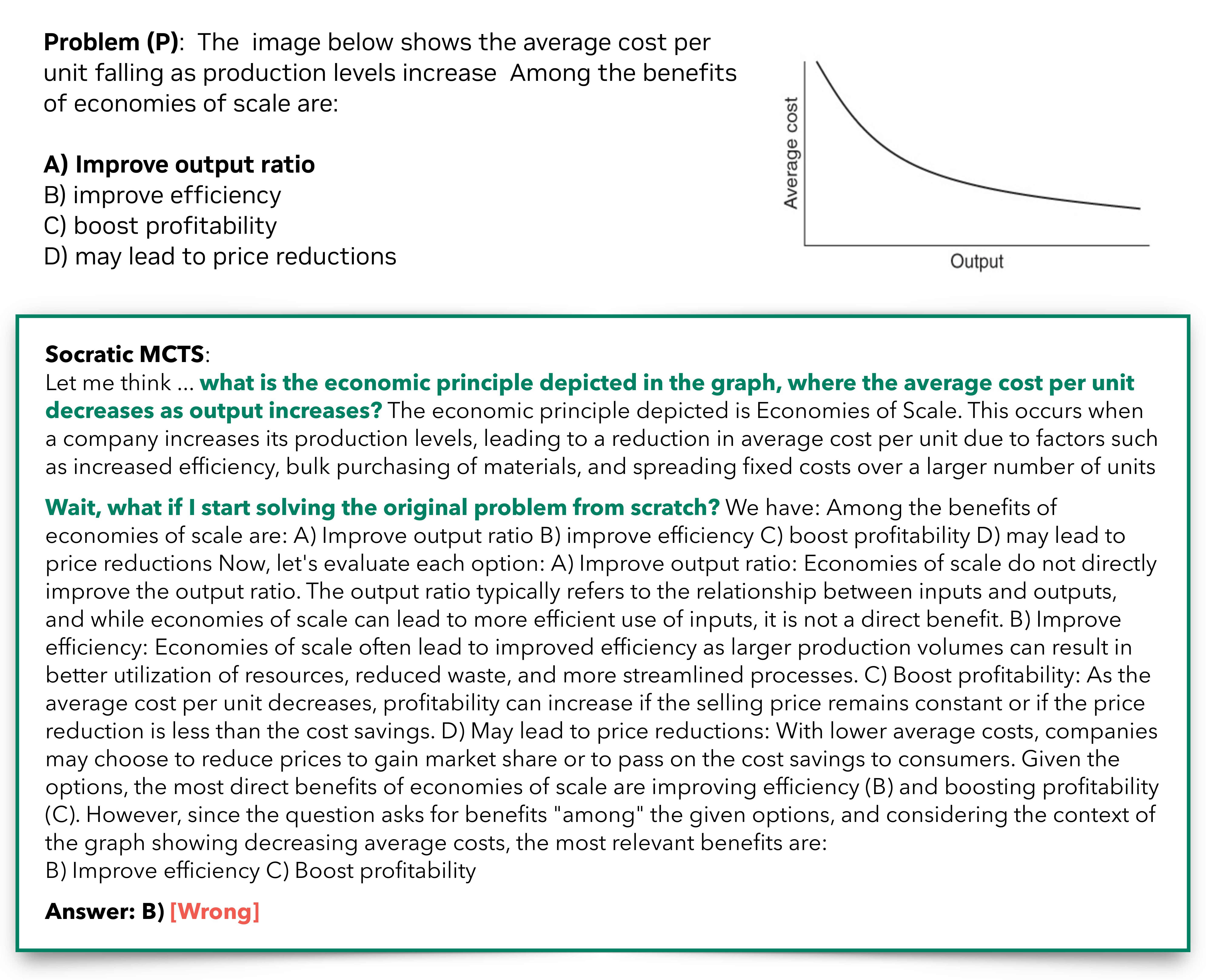}
    \label{fig:mmstar_fail}
}

\caption{\textbf{Failure cases of Socratic-MCTS} We show failure cases of Socratic-MCTS on MMStar and MMMU-Pro.}
\label{fig:mmmu-qual-comparison}
\end{figure*}
\section{Limitations}
While our approach demonstrates promising improvements, it comes with several limitations. First, non-autoregressive reasoning is inherently less GPU-efficient, making structured search methods like MCTS slower in practice. Without significant advances in compute efficiency, such structured approaches may remain less practical than simpler methods like majority voting with brute-force scaling.
Second, although internal agreement serves as a useful value signal in the absence of supervision, we observed that frozen VLMs tend to be overconfident in their outputs—often ignoring chain-of-thought cues, regardless of how they are preconditioned. Encouraging faithfulness of the CoT and output diversity in such models remains an open research challenge.
Finally, due to the computational cost of search, we did not explore hyperparameter tuning and multi-agent settings. 
These choices may offer further performance gains and warrant future investigation.

\bibliography{custom}
\clearpage
 \appendix
 \onecolumn
 
\section{Supplementary Material}

This appendix provides additional materials to support 
the main findings of the paper. 

    
    

\FloatBarrier

\section{Full text prompts and phrases used}\label{ref:app_prompts}

\begin{figure*}[h]
\centering   
\begin{tcolorbox}
\small
\begin{Verbatim}[breaklines=true, commandchars=\\\{\}]
starting_out_phrases = [
    "Let me think ...",
    "First, I need to consider ...",
]
\end{Verbatim}
\end{tcolorbox}
\caption{\textbf{\texttt{starting\_out\_phrases}: phrases used to begin the reasoning trajectory.}}\label{fig:starting-out-phrases}
\end{figure*}

\begin{figure*}[h]
\centering   
\begin{tcolorbox}
\small
\begin{Verbatim}[breaklines=true, commandchars=\\\{\}]
transition_phrases = [
    "Next, let me look at ...",
    "Moving on, I wonder ... ",
    "That leads me to the next point ...",
    "Expanding on that ...",
    "So what does this mean for ...",
    "Now, I need to think about ... "
]
\end{Verbatim}
\end{tcolorbox}
\caption{\textbf{\texttt{transition\_phrases}: phrases used to signal intermediate reasoning steps.}}\label{fig:transition-phrases}
\end{figure*}

\begin{figure*}[h]
\centering   
\begin{tcolorbox}
\small
\begin{Verbatim}[breaklines=false, commandchars=\\\{\}]
wrap_up_phrases = [
    "Summarizing the above and answering the original problem, we have:",
    "Wait, let me check if I made any mistakes above and answer the original problem:",
    "Okay, wrapping up any remaining calculations and answering the original problem, we have:",
    "Now, solving the original problem, we have:",
    "Um, let me summarize my thoughts and answer the original problem, we have:",
    "How do my previous thoughts connect to the original problem? Let me go back and solve it:",
    "Next, let me evaluate the above information recognizing that some of it may be incorrect:",
    "So, let me summarize the above information and directly answer the original problem:",
    "Wait, what if I start solving the original problem from scratch? We have:",
    "Wait, what if I start solving step-by-step the original problem from scratch? We have:",
    "Let's verify step-by-step the above information and answer the original problem:",
]
\end{Verbatim}
\end{tcolorbox}
\caption{\textbf{\texttt{wrap\_up\_phrases}: phrases used to conclude and produce the final answer.}}\label{fig:end-phrases}
\end{figure*}

\begin{figure*}[h]
\centering   
\begin{tcolorbox}
\small
\begin{Verbatim}[breaklines=true, commandchars=\\\{\}]
---
** Problem: $#$problem$#$
---
** Intermediate Reasoning: $#$partialreasoning$#$
---
Context: You have been provided with: 
* An image
* A problem related to that image
* Some intermediate reasoning steps that outline partial progress or initial thinking.
Your Goal:
- Formulate a single follow-up question that moves the reasoning process closer to solving the original problem.
Guidelines:
- Use the Socratic method by focusing on the next logical step in the problem-solving process.
Your question might:
- Clarify the core issue
- Challenge hidden assumptions
- Identify missing information or data
- Refine or expand upon a hypothesis
- Break the problem down into more manageable parts
- Probe deeper into the analysis or experimentation
- Prompt reevaluation or checking for errors
- Encourage synthesis to form a conclusion
- Spark reflection or consideration of next steps
Constraints:
- Ensure the question is directly relevant to the problem.
- Ensure the question is directly relevant to the intermediate reasoning steps.
- Ensure the question is asked in first-person perspective.
- Make the question open-ended enough to encourage deeper thinking, but specific enough to be actionable.
- If necessary use domain-specific terminology and aim to retrive domain-specific knowledge.
- Avoid simply restating the information already provided or present in the intermediate reasoning steps; instead, aim to advance the user's understanding or resolution of the problem.

Your response should be of the following format: 'Question: $Q' (without quotes) where Q is your proposed follow up question. - do not write anything other than 'Question: $Q' (without quotes) where Q is your proposed question."

Assistant: Question:
\end{Verbatim}
\end{tcolorbox}
\caption{\textbf{Text prompt used to generate follow up subquestions.}}\label{fig:prompt-description-to-mcqs}
\end{figure*}

\begin{figure*}[h]
\centering   
\begin{tcolorbox}
\small
\begin{Verbatim}[breaklines=true, commandchars=\\\{\}]
---
** Problem: $#$problem$#$
---
Context: You have been provided with: 
* An image
* A multiple choice question problem related to that image
Your Goal:
- Formulate a single follow-up question that moves the reasoning process closer to solving the original problem.
Guidelines:
- Use the Socratic method by focusing on the next logical step in the problem-solving process.
Your question might:
- Clarify the core issue
- Challenge hidden assumptions
- Clarify or ask to describe parts of the image that are relevant to solve the problem
- Identify missing information or data
- Refine or expand upon a hypothesis
- Break the problem down into more manageable parts
- Probe deeper into the analysis or experimentation
- Prompt reevaluation or checking for errors
- Encourage synthesis to form a conclusion
- Spark reflection or consideration of next steps
Constraints:
- Ensure the question is directly relevant to the problem.
- Ensure the quesiton is asked in first-person perspective.
- Ensure the question is directly relevant to the intermediate reasoning steps.
- Ensure the question is asked in first-person perspective.
- Make the question open-ended enough to encourage deeper thinking, but specific enough to be actionable.
- If necessary use domain-specific terminology and aim to retrive domain-specific knowledge.
- Avoid simply restating the information already provided or present in the intermediate reasoning steps; instead, aim to advance the user's understanding or resolution of the problem.
Your response should be of the following format: 'Question: $Q' (without quotes) where Q is your proposed follow up question. - do not write anything other than 'Question: $Q' (without quotes) where Q is your proposed question."
Assistant: Question:
\end{Verbatim}
\end{tcolorbox}
\caption{\textbf{Zero-Shot text prompt used to generate initial subquestions.}}\label{fig:prompt-description-to-mcqs}
\end{figure*}

\begin{figure*}[h]
\centering   
\begin{tcolorbox}
\small
\begin{Verbatim}[breaklines=true, commandchars=\\\{\}]
$#$problem$#$
Answer with the option letter from the given choices directly. The last line of your response should be of the following format: 'Answer: $LETTER' (without quotes) where LETTER is one of options.
\end{Verbatim}
\end{tcolorbox}
\caption{\textbf{Text prompt from~\cite{mmmu} used to evaluate Direct.}}\label{fig:prompt-description-to-mcqs}
\end{figure*}

\begin{figure*}[h]
\centering   
\begin{tcolorbox}
\small
\begin{Verbatim}[breaklines=true, commandchars=\\\{\}]
$#$problem$#$
Answer the preceding multiple choice question. The last line of your response should be of the following format: 'Answer: $LETTER' (without quotes) where LETTER is one of options. Think step by step before answering.
\end{Verbatim}
\end{tcolorbox}
\caption{\textbf{Text prompt from~\cite{mmmu} used to evaluate zero-shot CoT.}}\label{fig:prompt-description-to-mcqs}
\end{figure*}


\end{document}